\documentclass[lettersize,journal]{IEEEtran}
\usepackage{amsmath,amsfonts}
\usepackage{algorithmic}
\usepackage{algorithm}
\usepackage{array}
\usepackage[caption=false,font=normalsize,labelfont=sf,textfont=sf]{subfig}
\usepackage{textcomp}
\usepackage{stfloats}
\usepackage{url}
\usepackage{verbatim}
\usepackage{graphicx}
\usepackage{cite}
\usepackage{bm}
\usepackage{bbding}
\usepackage{hyperref}
\usepackage{tabularray}
\usepackage{multirow}
\usepackage{colortbl}
\usepackage{booktabs}
\usepackage{hhline}
\usepackage{amsmath}
\usepackage{pifont}
\usepackage{wrapfig}
\usepackage{amssymb}
\usepackage{multirow}

\usepackage{cleveref}

\definecolor{darkblue}{rgb}{0, 0.12, 0.55}
\definecolor{darkgreen}{rgb}{0, 0.55, 0.12}
\definecolor{darkred}{rgb}{0.6,0,0}
\definecolor{darkgreen}{rgb}{0,0.6,0}
\definecolor{gray}{gray}{0.9}
\definecolor{lightblue}{rgb}{0.68, 0.85, 0.9}
\definecolor{cvprblue}{rgb}{0.21,0.49,0.74}
\newcolumntype{H}{>{\setbox0=\hbox\bgroup}c<{\egroup}@{}}

\usepackage{soul}

\begin{document}
\IEEEpubid{\begin{minipage}{\textwidth}\ \\[40pt]
  \centering\footnotesize
  Copyright \copyright\ 2024 IEEE. Personal use of this material is permitted. However, permission to use this material for any other purposes must be obtained from the IEEE by sending an email to pubs-permissions@ieee.org.
\end{minipage}}

\title{Generalization Boosted Adapter \\ for Open-Vocabulary Segmentation}

\author{Wenhao Xu, 
Changwei Wang,
Xuxiang Feng,
Rongtao Xu$^{*}$, 
Longzhao Huang,
Zherui Zhang,
Li Guo,
\\
Shibiao Xu$^{*}$,~\IEEEmembership{Member,~IEEE}

\thanks{*Rongtao Xu and Shibiao Xu are the corresponding authors (xurongtao2019@ia.ac.cn; shibiaoxu@bupt.edu.cn).
}
\thanks{Wenhao Xu, Shibiao Xu, Longzhao Huang, Zherui Zhang and Li Guo are with School of Artificial Intelligence, Beijing University of Posts and Telecommunications, Beijing 100876, China. 
Changwei Wang is with the Key Laboratory of Computing Power Network and Information Security, Ministry of Education, Shandong Computer Science Center (National Supercomputer Center in Jinan), Qilu University of Technology (Shandong Academy of Sciences), Jinan, 250013, China; Shandong Provincial Key Laboratory of Computing Power Internet and Service Computing, Shandong Fundamental Research Center for Computer Science, Jinan, China.
Xuxiang Feng is with the Aerospace Information Research Institute, Chinese Academy of Sciences, China.
Rongtao Xu is with the State Key Laboratory of Multimodal Artificial Intelligence Systems, Institute of Automation, Chinese Academy of Sciences, Beijing, China
}
\thanks{This work is supported by Beijing Natural Science Foundation No. JQ23014, and by the Science and Disruptive Technology Program, AIRCAS (No. E2Z218020F), and by the National Natural Science Foundation of China (Nos. 62271074, 62071157, 62302052, 62171321 and 62162044).}
}

\markboth{IEEE Transactions on Circuits and Systems for Video Technology,~Vol.~xx, No.~x, July~2024}%
{Shell \MakeLowercase{\textit{et al.}}: A Sample Article Using IEEEtran.cls for IEEE Journals}


\maketitle

\begin{abstract}

Vision-language models (VLMs) have demonstrated remarkable open-vocabulary object recognition capabilities, motivating their adaptation for dense prediction tasks like segmentation. However, directly applying VLMs to such tasks remains challenging due to their lack of pixel-level granularity and the limited data available for fine-tuning, leading to overfitting and poor generalization. To address these limitations, we propose Generalization Boosted Adapter (GBA), a novel adapter strategy that enhances the generalization and robustness of VLMs for open-vocabulary segmentation.
GBA comprises two core components: (1) a Style Diversification Adapter (SDA) that decouples features into amplitude and phase components, operating solely on the amplitude to enrich the feature space representation while preserving semantic consistency; and (2) a Correlation Constraint Adapter (CCA) that employs cross-attention to establish tighter semantic associations between text categories and target regions, suppressing irrelevant low-frequency ``noise'' information and avoiding erroneous associations.
Through the synergistic effect of the shallow SDA and the deep CCA, GBA effectively alleviates overfitting issues and enhances the semantic relevance of feature representations. As a simple, efficient, and plug-and-play component, GBA can be flexibly integrated into various CLIP-based methods, demonstrating broad applicability and achieving state-of-the-art performance on multiple open-vocabulary segmentation benchmarks. 
\end{abstract}


\section{Introduction}
Image segmentation is one of the most classic and fundamental problems in computer vision~\cite{xu2023scd,xu2023wave,xu2024skinformer} and autonomous driving~\cite{wang2024iterative,wang2023coordination,xu2024mrftrans,dalf}, aiming to assign a semantic category to every pixel. Modern segmentation methods\cite{cheng2022masked,xu2021dc,dannet} rely heavily on large-scale annotated data, but typical datasets contain only tens to hundreds of categories. The high cost of data collection and annotation limits the application of these methods in practical scenarios that require handling open-vocabulary objects.

\begin{figure}[!ht]
    \centering
    \includegraphics[width=8cm, height=6cm]{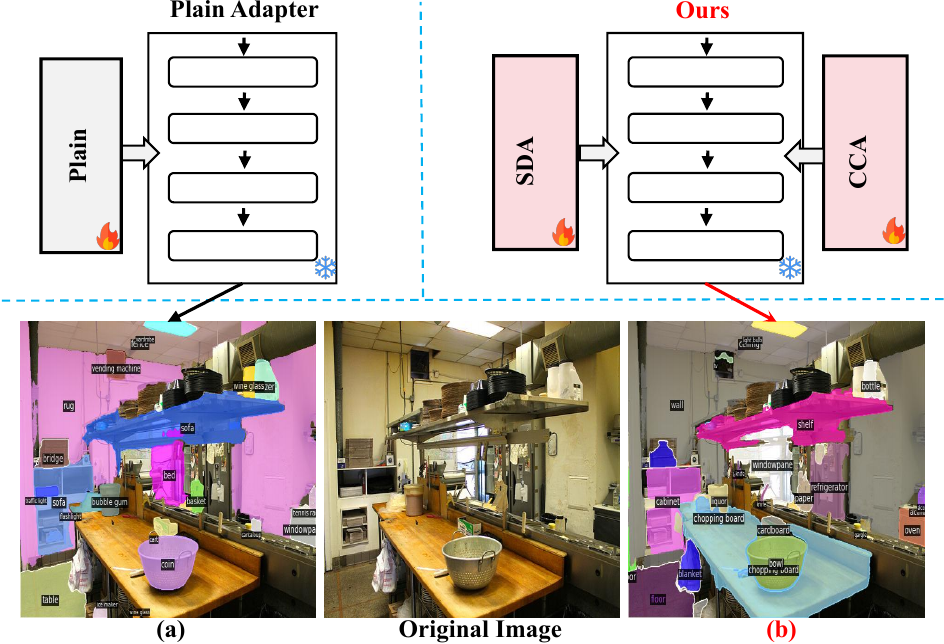}
    \caption{ We use FC-CLIP as the baseline to investigate the performance of different adapter methods in dense prediction tasks. As shown in \textbf{(a)}, the model with plain adapter incorrectly identifies the object as a `sofa' and further misclassifies the `wall' as a "rug" due to false associations. In contrast, our proposed GBA method accurately recognizes the `shelf' and `wall' in the image, as illustrated in \textbf{(b)}. This improvement can be attributed to the GBA's ability to enhance the model's generalization capability and suppress false associations for dense prediction tasks.
    }

    \label{fig:intro}
\end{figure}

Recently, vision-language models (VLMs)\cite{ALIGN, CLIP} have gained significant attention due to their remarkable open-vocabulary object recognition capability. This tremendous success has motivated us to explore their adaptability to the segmentation task. VLMs have demonstrated outstanding feature representation capabilities through cross-modal contrastive learning. However, such representation lacks pixel-level granularity, making it challenging to directly apply to dense prediction tasks.
Inspired by work in natural language processing\cite{jiang2020can,shin2020autoprompt}, a series of methods have been proposed in the vision domain, aiming to efficiently adapt VLMs to open-vocabulary segmentation tasks. Existing methods include prompt learning\cite{lu2022prompt,xuspt} and refining learned representations using adapters\cite{zhang2021tip,gao2024clip}.
Despite the progress made by prompt tuning and adapter techniques in open-vocabulary tasks, fine-tuning approaches that introduce a small number of learnable parameters for downstream tasks still have limitations due to the significantly smaller size of the fine-tuning dataset compared to the pre-training dataset used for VLMs:
(1) Fine-tuning models with limited data may lead to overfitting on specific patterns in the training samples, making it difficult to effectively learn general visual concepts, resulting in overfitting or poor generalization performance\cite{zhou2022coop,ma2022understanding}. 
(2) Training data contains a large amount of ``noise'' information irrelevant to semantic categories, such as background textures and object styles\cite{Chang2023LearningSR,song2023FD}. Direct fine-tuning may cause the model to overly focus on this irrelevant information and establish "false associations" with the correct categories, undermining the original robustness and generalization capabilities of VLMs. 

These methods are limited in their performance on open-vocabulary segmentation tasks because they cannot significantly alter the fundamental visual representations encoded in the model (e.g., CLIP), often leading to overfitting to the training data or poor out-of-distribution generalization. As illustrated in Fig.\ref{fig:intro} (a), the model with a plian adapter incorrectly recognizes a shelf, consequently misidentifying other objects as well.
In this paper, we propose introducing adapter modules on top of the CLIP backbone network to enhance CLIP's generalization and robustness in open-vocabulary segmentation tasks, as shown in Fig.\ref{fig:intro} (b). To achieve this objective, we devise a novel adapter strategy, termed \textbf{Generalization Boosted Adapter} (GBA), which comprises two core components:
First, we introduce a \textbf{Style Diversification Adapter (SDA)} that decouples features into amplitude and phase components through Fourier transform, corresponding to style and content information, respectively. We operate solely on the amplitude component, aiming to alter the style while preserving semantic consistency, thereby enriching the feature space representation of the visual encoder.
Second, we incorporate a \textbf{Correlation Constraint Adapter (CCA)} into the deeper layers of the visual encoder. This adapter employs a cross-attention mechanism to establish a tighter semantic association between text categories and target regions in the image, guiding the suppression of irrelevant low-frequency "noise" information. Consequently, it avoids erroneous associations between correct categories and irrelevant information, thus enhancing category matching accuracy.
Moreover, GBA serves as a simple and efficient modular component that can be flexibly integrated into various CLIP-based methods, rendering it a universal solution for addressing diverse open-vocabulary downstream tasks.
Through the synergistic effect of the shallow SDA and the deep CCA, the proposed GBA method effectively alleviates overfitting issues and enhances the semantic relevance of feature representations, thereby significantly boosting the performance of open-vocabulary segmentation tasks. As a simple, efficient, and plug-and-play component, GBA demonstrates broad applicability and can be readily employed in various CLIP-based open-vocabulary downstream tasks.
Our paper makes the following contributions:

\textbf{(i)} To mitigate the overfitting issue caused by limited fine-tuning data, we propose a \textbf{Style Diversification Adapter} (SDA) to enhance feature diversity, preventing the model from overly memorizing specific patterns in the training data and improving generalization.
while maintaining content invariance through a content consistency loss.  

\textbf{(ii)} To address the issue of visual feature representation quality, We propose a \textbf{Correlation Constraint Adapter} (CCA) that implicitly enforces a false association constraint by suppressing irrelevant low-frequency information, thereby enhancing the semantic relevance between text categories and downstream images, which greatly benefits the category matching process.

\textbf{(iii)} The proposed GBA method leverages the synergistic effect of two different adapter strategies in a plug-and-play manner, achieving state-of-the-art results on multiple benchmarks.

\section{Related Works}
\subsection{Vision Language Models}

Vision-language models aim to learn generic representations that align visual and textual information. Early works, such as UNITER \cite{chen2020uniter} and Oscar \cite{li2020oscar}, employed a two-stage approach: first, they pretrained object detectors on smaller datasets to extract semantic representations, and then fine-tuned these representations on downstream tasks like visual question answering (VQA) and image captioning. However, these methods required carefully designed fusion encoders to integrate cross-modal interactions.
Inspired by the success of pretraining in computer vision (CV) and natural language processing (NLP), many researchers have proposed approaches to pretrain large-scale models that process both visual and language modalities simultaneously. A typical visual-language model consists of four key components: a visual encoder, a language encoder, a fusion encoder, and loss functions. Building on the success of foundation models in CV and NLP domains, such as BERT \cite{devlin2018bert} and Vision Transformers \cite{dosovitskiy2020image, li2020learning}, the multimodal learning community has leveraged these large-scale foundation models to boost performance \cite{zhang2024navid,zhang2023task}. For example, VisualBERT \cite{li2019visualbert}, OSCAR \cite{li2020oscar}, and Uniter \cite{chen2020uniter} utilize BERT \cite{devlin2018bert} to preprocess raw texts and have achieved impressive results on multimodal tasks like VQA.
Recent breakthroughs in large-scale visual-language models, such as CLIP \cite{CLIP} and ALIGN \cite{ALIGN}, have demonstrated that dual-encoder models pretrained on large-scale image-text pair datasets can learn representations with cross-modal alignment. These models have rapidly advanced zero-shot and open-vocabulary downstream tasks.\cite{CLIP} and \cite{ALIGN} have shown that visual-language contrastive learning can generate transferable features for downstream tasks, and the multimodal interaction can be well explained by simply computing the dot product between visual and language embeddings. Recently, some studies have applied large-scale visual-language models to open-vocabulary  segmentation and other downstream tasks, achieving impressive results and confirming the benefits of pretraining on large-scale text-image pairs.
Inspired by these methods, the current work explores the introduction of specially designed dual adapters, allowing models to produce superior open-vocabulary classification segmentation results while preserving the cross-modal alignment capabilities of large-scale vision-language models. 

\subsection{Adapting Vision-Language Models}
The concept of adapters, initially introduced in the field of natural language processing (NLP) for fine-tuning large pre-trained models on downstream tasks, has recently garnered significant attention in the vision-language domain. Large-scale pre-trained language models have demonstrated remarkable success in various natural language tasks, such as question answering, sentence completion, and language translation\cite{radford2019language,Guu2020REALMRL}. However, due to the immense scale of these models, training or fine-tuning them is prohibitively expensive. To address this issue, researchers have proposed several efficient adaptation methods, including designing lightweight adapter modules, bias adjustment calibration, or learning task-specific soft text prompts, enabling scholars to leverage the power and generality of large pre-trained language models on downstream tasks of interest.
In the vision-language domain, methods like CLIP\cite{CLIP} learn the embedded representations of images and text through contrastive learning, where the embeddings are encoded by independent image and language encoders. Essentially, CLIP\cite{CLIP} learns the representations of both modalities by pulling the image representations closer to their paired text representations while pushing away the text embeddings corresponding to different images. The advantage of these methods lies in their ability to learn from vast amounts of unstructured image and text data available on the internet, allowing vision-language models to be applied to various downstream tasks in zero-shot or few-shot settings. 
Inspired by the adaptation of natural language models, two primary approaches have been proposed to adapt CLIP-like models to downstream vision tasks: prompt learning and feature adaptation. Prompt learning methods, such as CoOp\cite{zhou2022coop} and CoCoOp\cite{zhou2022conditional}, aim to automatically construct text prompts by optimizing learnable vectors. In contrast, feature adaptation methods, such as CLIP-Adapter\cite{gao2021clip} and Tip-Adapter\cite{zhang2021tip}, directly adjust the representations extracted from the visual and text encoders of CLIP-like models. CLIP-Adapter\cite{gao2021clip} adds a lightweight fully-connected neural network adapter applied to frozen CLIP\cite{CLIP} features and fine-tunes the parameters on the downstream task of interest with limited supervision. Tip-Adapter achieves better results by constructing a key-value cache model from few-shot examples and performing fewer fine-tuning steps. Furthermore, a tune-free version of Tip-Adapter\cite{zhang2021tip} has been proposed, which adapts faster during training but yields inferior performance. Although these adapter methods are relatively efficient, they are insufficient for making large-scale changes to the fundamental representations extracted from the backbone visual or text encoders, which may pose a problem when the evaluation task differs significantly from the distribution encountered during training.
Recently, several works have attempted to add adapters to visual foundation models to better perform pixel-level prediction tasks. SAM-Adapter\cite{chen2023sam} adapts SAM to specific scenarios, such as shadow detection and camouflaged object detection, by adding learnable adapters. SAN\cite{san} improves the performance of open-vocabulary semantic segmentation by adding two parallel side networks as adapters on top of a frozen backbone network.
Despite the proven benefits of the aforementioned basic adapters in adapting pre-trained models to downstream tasks, they still suffer from overfitting issues due to the relatively small size of the downstream datasets used for fine-tuning. Based on this premise, we have specifically designed novel adapter strategies to address the problems of feature robustness and spurious correlations that arise when adapting to downstream segmentation tasks, stemming from the inherent limitations of CLIP\cite{CLIP} and the small scale of downstream datasets. 

\subsection{Open Vocabulary Segmentation}
Deep learning \cite{xu2022instance, wang2022cndesc, xu2024local, xu2023domainfeat,ji2020encoder} 
 and image segmentation have recently achieved remarkable success\cite{ xu2023rssformer,xu2023dual,xu2024hcf,wang2023treating,chen2023adversarial,liu2020guided}.
Open vocabulary segmentation aims to segment target categories that are unseen during training, which remains a challenging task due to the limited availability of annotated data and the vast diversity of objects in the real world.
The existing approaches can be divided into two aspects: mapping visual features into semantic space and cross-modal alignment with pre-trained models \cite{ ding2022decoupling, xu2021simple}. 
For the mapping aspect, SPNet\cite{xian2019semantic} encodes visual features to the semantic embedding space and then projects each pixel feature to predict probabilistic outcomes through a fixed semantic word encoding matrix.
ZS3Net\cite{bucher2019zero}  generates the pixel-level features of unseen classes in the semantic embedding space and adopts the generated features to supervise a visual segmentation model. 
STRICT\cite{pastore2021closer} introduces a self-training technique into SPNet to improve the segmentation performance of unseen classes. 
Recent advancements in large-scale visual language modeling have significantly contributed to the progress of open-vocabulary semantic segmentation. 
LSeg\cite{li2022language} learns a CNN model to compute per-pixel image features to match with the text embeddings embedded by the pre-trained text model. 
ZegFormer\cite{ding2022decoupling} and ZSSeg\cite{xu2021simple} leverage the visual model to generate the class-agnostic masks, and use the pre-trained text encoder to retrieve the unseen class masks. 
XPM\cite{huynh2022open} utilizes the region-level features to match CLIP-based text embeddings to accomplish the open vocabulary instance segmentation. 
Some of these studies\cite{liang2023open,zhou2022extract} fine-tuned the visual language pre-training model, but would compromise the cross-modal alignment ability of the visual language model.
However, fine-tuning the visual language pre-training model, as done in some studies\cite{liang2023open,zhou2022extract}, may compromise the cross-modal alignment ability of the model.

To address this issue, various approaches have been proposed. SimSeg\cite{SimSeg} and MaskCLIP\cite{MaskCLIP} introduce a two-phase framework that first generates class-independent masks and then assigns classes to the masks using a frozen CLIP. ODISE\cite{ODISE} and FreeSeg\cite{FreeSeg} enhance the quality of open-vocabulary semantic segmentation by incorporating a diffusion model and a multi-granularity concepts encoder, respectively. FC-CLIP\cite{fc-clip} utilizes a shared frozen convolutional CLIP backbone within an efficient single-stage framework. SAN\cite{san} extends the frozen CLIP with two side branches: one for predicting mask proposals and another for predicting attentional bias, which is used to identify mask classes.
Building upon these advancements, our work further investigates the adaptability of the frozen-parameter CLIP model to the open-vocabulary semantic segmentation task. We propose leveraging a novel dual adapter strategy to enrich the feature representation space and alleviate spurious associations between categories and noise, thereby enhancing the generalization performance of open-vocabulary semantic segmentation.

\section{Method}
\subsection{Task Definition}
Given an image $\mathbf{I}\in\mathbb{R}^{H\times W\times 3}$, where $H$ and $W$ denote the height and width of the image respectively, open-vocabulary semantic segmentation aims to segment the image into a set of masks with associated semantic classes:

\begin{equation}
\mathcal{Y} = {(\mathcal{M}_{i}, \mathcal{C}_{i})}_{i=1}^{N} .
\end{equation}
where $\mathcal{M}_{i} \in {{0,1}}^{H \times W}$ represents the ground truth mask and $\mathcal{C}_{i} \in \mathcal{S}_{train} \bigcup \mathcal{S}_{test}$ denotes the corresponding ground truth class label.

Open-vocabulary semantic segmentation is more challenging than traditional image segmentation tasks\cite{he2017mask,kirillov2019panoptic} because the inference classes are not observed during training.
During evaluation, the test categories $\mathbf{C}_{\text{test}}$ are different from $\mathbf{C}_{\text{train}}$, containing novel categories not seen in training, i.e., $\mathbf{C}_{\text{train}} \neq \mathbf{C}_{\text{test}}$.

\subsection{Baseline.}
\begin{figure*}[ht]
    \centering
    \includegraphics[width=16cm, height=7cm]{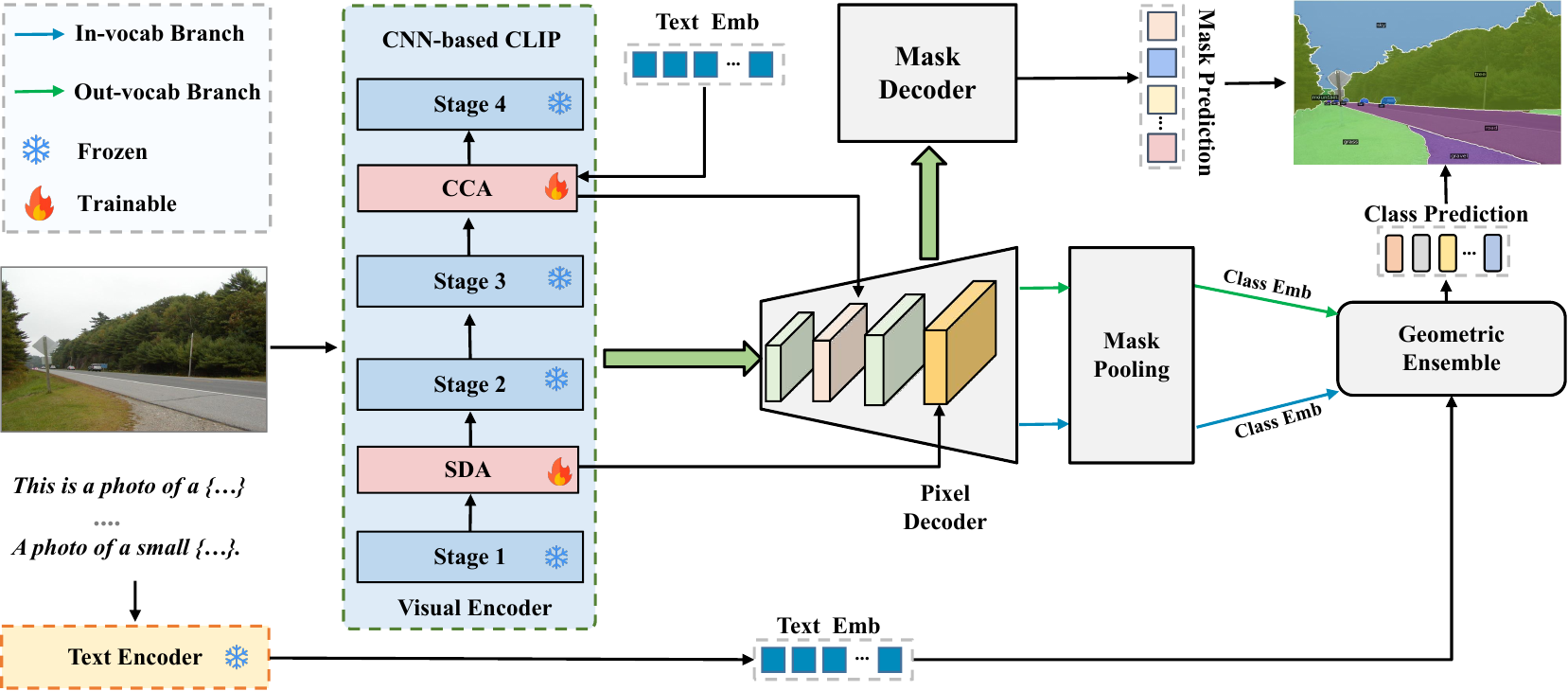}
    \caption{\textbf{Overview of the GBA framework.}
 Single-stage open-vocabulary  segmentation methods akin to FC-CLIP comprise three key components: a mask generator, an in-vocabulary classifier, and an out-of-vocabulary classifier. All these components are constructed based on features extracted from a frozen CNN-based CLIP backbone that employs the proposed two learning feature augmentation adapters, namely, the Style Diversification Adapter (SDA) and the Correlation Constraint Adapter (CCA).
    }
    \label{fig:overview}
\end{figure*}
We adopt the FC-CLIP\cite{fc-clip} as our baseline and integrate our proposed adapters as lightweight components.
The image encoder is configured as a ConvNext-Large model, comprising four stages. Each stage contains a different number of blocks: 3 in the first, 3 in the second, 27 in the third, and 3 in the fourth.
In contrast, the text encoder is structured as a 16-layer transformer, each layer being 768 units wide and featuring 12 attention heads. We harness the power of multi-scale features extracted by the image encoder. These features are represented as feature maps of varying widths and scales: a 192-wide feature map downscaled by a factor of 4, a 384-wide map downscaled by 8, a 768-wide map downscaled by 16, and a 1536-wide map downscaled by 32.
FC-CLIP incorporates a frozen-parameter convolutional CLIP backbone (ConvNeXt\cite{liu2022convnet}) into the Mask2Former\cite{cheng2022masked} segmentation pipeline, achieving a simple yet effective single-stage open-vocabulary semantic segmentation approach.

\subsection{Generalization Boosted Adapter: Diversifying Features and Correlation Constraint}

CLIP is pre-trained through image-level contrastive learning, and thus using the frozen CLIP visual encoder to extract features for downstream dense prediction tasks leads to suboptimal results. This is because CLIP utilizes image-level supervision signals during pre-training, failing to capture pixel-level detail features. Moreover, when adapting CLIP features to dense prediction tasks, their spatial attention is commonly scattered across different regions of the image, resulting in spurious correlations when fine-tuning the learnable parameters on limited data.
To alleviate the above issues, a common approach is to insert learnable adapters into CLIP. Traditional adapter techniques primarily consist of linear layers, downsampling layers, nonlinear activation layers, upsampling layers, and skip connections. However, due to the small number of parameters in these standard adapters, they are prone to overfitting during training, excelling at handling seen classes but lacking generalization capability for unseen open-vocabulary classes.
Based on these observations, we propose the Generalization Boosted Adapter (GBA) dual-adapter design paradigm, as shown in Fig.\ref{fig:overview}. GBA comprises two adapter modules with distinct functionalities, respectively enhancing spatial feature style diversity adaptability and suppressing spurious correlations, aiming to improve the model's generalization ability for open-vocabulary semantic segmentation.

In our proposed lightweight plain adapter architecture, we introduce multiple convolutional layers with varying receptive fields to enhance the feature extraction capability of the adapter. Specifically, the feature maps are sequentially processed by three convolutional layers with kernel sizes of 3×3, 5×5, and 7×7, respectively. We compute the average of these three layers and employ a 1×1 convolution to aggregate the features. After applying a SiLU layer for non-linear activation, the feature maps undergo a feature augmentation strategy module. Additionally, we incorporate residual connections. Finally, a simple projection layer yields the output feature maps.

\begin{figure}[!ht]
    \centering
    \includegraphics[width=6.6cm, height=4cm]{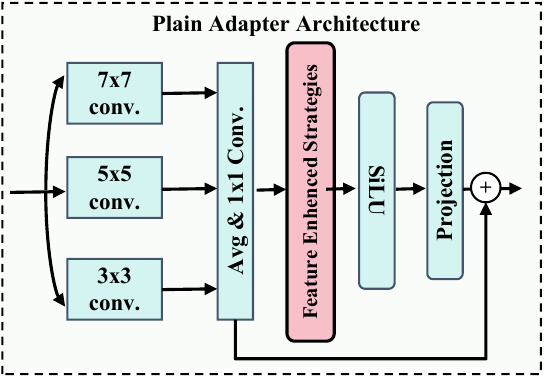}
    \caption{ \textbf{The detail of plain adapter}. Adapters exhibiting diverse generalization capacities can be realized through the utilization of varied feature enhenced strategies.
    }
    \label{fig:plain}
\end{figure}

As illustrated in Fig.~\ref{fig:plain}, our proposed two adapter modules adopt a similar base architecture but employ different feature augmentation strategies. In the following sections, we will elaborate on the design details and working mechanisms of these two strategies.

\subsubsection{Style Diversification Adapter}

Extending feature-level style variations has been recognized as an effective approach to enhance model robustness. Numerous studies, such as mixstyle and AdaIN, have been inspired by the observation that the statistical feature distributions in convolutional neural networks (CNNs) contain crucial style information. These methods synthesize stylized features by adjusting feature statistics, thereby increasing feature diversity. Well-known normalization techniques, including normalization (BN)\cite{Ioffe2015BatchNA}, layer normalization (LN)\cite{ba2016layer}, and instance normalization (IN)\cite{Ulyanov2016InstanceNT}, are commonly employed in these approaches since normalization statistics encapsulate style information, and normalization can effectively extract style from features.
However, when altering or removing style, semantic content may also undergo changes\cite{Jin2020StyleNA,ibn}. Recent research\cite{Chen2021AmplitudePhaseRR,Lee2023DecomposeAC}. has discovered that utilizing Fourier transform can effectively decompose input images into amplitude and phase components, which represent style and content, respectively. 
In tackling the single-domain generalization challenge, Wang~\cite{Wang2023FFMIO} leverages feature frequency modulation to synthesize images with diverse styles, effectively bolstering the generation of hard samples.
Based on the aforementioned insights, we propose a novel frequency-domain decomposition-based feature diversification strategy. This strategy aims to enrich style diversity while simultaneously preserving semantic integrity. By mapping features to the frequency domain and manipulating amplitude and phase components independently, we can adjust style and content separately, thereby generating rich style variations while maintaining semantic content consistency, as shown in Fig.\ref{fig:adapter} (a).

First, we extract the base style features of the input $\mathbf{x}$ and perform frequency decomposition. We compute the mean $\mu_{base}$ and standard deviation $\sigma_{base}$ of the base style features of $\mathbf{x}$, where $C$ denotes the number of channels, and $H$ and $W$ denote the height and width, respectively. Simultaneously, we apply the Fast Fourier Transform (FFT) to the original sample $x \in \mathbb{R}^{C \times H \times W}$ to obtain the frequency domain features $F(x)$, which we then decompose into the amplitude $ \mathbf{a}$ and phase $ \mathbf{p}$:

\begin{equation}
     \mathbf{a} = \sqrt{F(x)_{\text{real}}^2 +F(x)_{\text{img}}^2} ,
\end{equation}
\begin{equation}
     \quad \mathbf{p} = \arctan\left(\frac{F(x)_{\text{img}}}{F(x)_{\text{real}}}\right) ,
\end{equation}  

Here, $F_{\text{real}}$ and $F_{\text{img}}$ represent the real and imaginary parts of the Fourier coefficients, respectively. Next, we perform normalization and style fusion. 
 
We then element-wise multiply $\mu_{base}$ and $\sigma_{base}$ with the combination weights $\mathbf{W} \in \mathbb{R}^{C}$ sampled from the Dirichlet distribution $B(\alpha_1, \ldots, \alpha_C)$ to generate the amplitude features of the new style:

\begin{equation}
    \mathbf{\mu} = \mathbf{W}\cdot \mu_{base}, \qquad \sigma = \mathbf{W} \cdot \sigma_{base} ,
\end{equation}
\begin{equation}
\centering
     \mathbf{a}_{\text{new}} = \sigma \cdot  \mathbf{a} +\mu ,
\end{equation}

Finally, we recombine the new style amplitude $\mathbf{a}_{\text{new}}$ with the phase $\mathbf{p}$ of the original sample and perform the Inverse Fast Fourier Transform (IFFT) to generate the sample $\tilde{\mathbf{x}} \in \mathbb{R}^{C \times H \times W}$ with the new style:

\begin{equation}
     \tilde{\mathbf{x}} = \text{IFFT}\left(\text{Compose}(\mathbf{a}_{\text{new}}, \mathbf{p}\right)) .
\end{equation}  
Throughout this process, by keeping the phase of the original sample unchanged, we ensure content consistency and achieve the goal of preserving content information during style transformation. Meanwhile, by normalizing the amplitude features and fusing different styles, we generate samples with new styles.

\subsubsection{Correlation Constraint Adapter}

\begin{figure}[!ht]
    \centering
    \includegraphics[width=8cm, height=6.5cm]{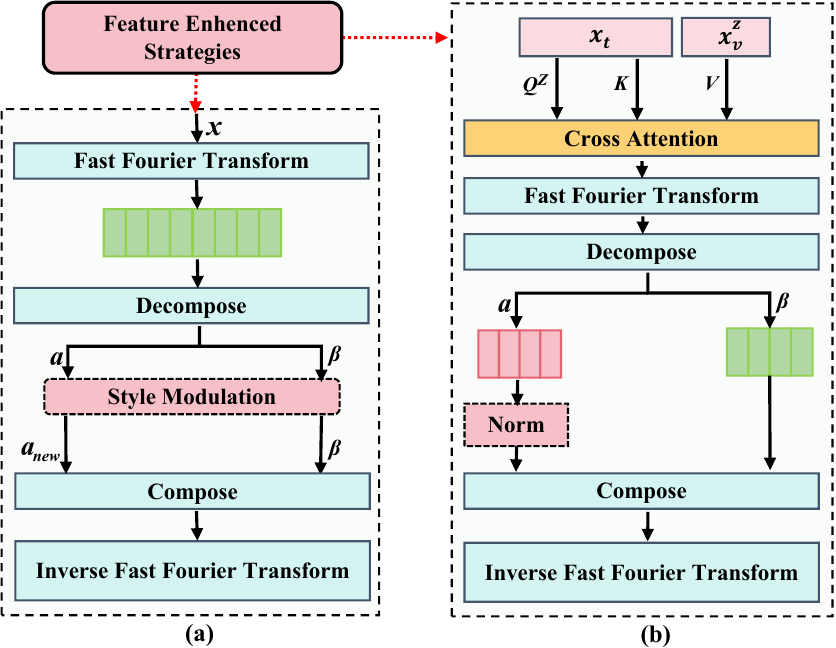}
    \caption{
 The details of Style Diversification Adapter (SDA) (a) and Correlation Constraint Adapter (CCA) (b).}
    \label{fig:adapter}
\end{figure}
False correlation, often caused by background elements irrelevant to the target label, is a common issue in visual recognition tasks. Previous research\cite{bai2022improving,sun2005background} has revealed that high-frequency components, such as object edges and contours, exhibit stronger associations with semantic features compared to the relatively lower frequencies of backgrounds and object surfaces\cite{liu2009background}. Inspired by this insight, we propose a novel approach to mitigate the impact of false correlation on model performance, as shown in Fig.\ref{fig:adapter} (b).
By emphasizing learning from high-frequency components and mitigating the risk of learning erroneous correlations, the model can more effectively capture the semantic features of objects. To achieve this goal, we introduce a cross-attention mechanism for semantic interaction to model the relevance between text embeddings and multi-scale visual features, thereby guiding the model to learn semantically-relevant high-frequency information.
Specifically, the semantic interaction can be formulated as:

\begin{equation}
Attn(\mathbf{Q}^{z}, \mathbf{K},  \mathbf{V}) = \text{softmax}(\frac{\mathbf{Q}^{z}\mathbf{K}^T}{\sqrt{d_k}}) \mathbf{V} \text{,}
\end{equation}
\begin{equation}
\mathbf{Q}^{z} = \phi_q (\mathbf{X}_{v}^z)\text{,} \quad  \mathbf{K} = \phi_k (\mathbf{X}_{t})\text{,} \quad  \mathbf{V} = \phi_v (\mathbf{X}_{t}) \text{,}
\end{equation}
where $\mathbf{X}_{v}^z$ denotes the visual features from the $z$-th layer of the decoder in the masked extractor, and $\mathbf{X}_{t}$ represents the text category features.
$\mathbf{Q}^{z}, \mathbf{K}, \mathbf{V}$ represent the query, key, and value embeddings generated by the projection layers $\phi_q, \phi_k, \phi_v$, respectively. $\sqrt{d_k}$ is the scaling factor. The attention relation is then utilized to enhance the visual features:

\begin{equation}
  \hat{\mathbf{X}_{v}^z} = \mathcal{L} \{Attn[\phi_q (\mathbf{X}_{v}^{z}),\phi_k (\mathbf{X}_{t}), \phi_v (\mathbf{X}_{t})]\} \text{,}
  \end{equation}
where $\mathcal{L}$ represents the output projection layer. 
The enhanced visual features $\hat{\mathbf{X}_{v}^z}$ facilitate emphasizing the visual information relevant to the given text category, thereby better capturing the high-frequency semantic features of objects.

Subsequently, we perform operations in the Fourier frequency domain to facilitate the model's utilization of high-frequency information. By operating in the frequency domain, we can exert finer-grained control over the model's attention to different frequency components, thereby more effectively mitigating the impact of error correlation.
Specifically, we first apply the Fast Fourier Transform (FFT) to the visual features $\hat{\mathbf{X}_{v}^z} \in \mathbb{R}^{C \times H \times W}$, mapping them into the frequency domain to obtain the frequency domain features $F(\hat{\mathbf{X}_{v}^z})$. We then decompose $F(\hat{\mathbf{X}_{v}^z})$ into the amplitude component $\mathbf{a}$ and the phase component $\mathbf{p}$. The phase component $\mathbf{p}$ primarily encodes high-frequency details such as edges and textures in the image, while the amplitude component $\mathbf{a}$ predominantly reflects low-frequency global characteristics such as smoothness and contrast.
To enhance the influence of high-frequency structural information, we perform normalization on the amplitude component $a$:

\begin{equation}
\mathbf{a}_{\text{norm}} = \frac{\mathbf{a} - \mu(\mathbf{a})}{\sigma(\mathbf{a})},
\end{equation}
where $\mu(\mathbf{a})$ and $\sigma(\mathbf{a})$ represent the mean and standard deviation of $a$, respectively. Through this normalization operation, we can implicitly amplify the relative importance of high-frequency structural information while suppressing the impact of low-frequency global features.
Finally, we recombine the normalized amplitude $\mathbf{a}_{\text{norm}}$ with the phase $\mathbf{p}$ of the original visual features and perform the Inverse Fast Fourier Transform (IFFT) to generate the processed visual features $\hat{\mathbf{X}} \in \mathbb{R}^{C \times H \times W}$:
\begin{equation}
    \hat{\mathbf{X}} = \text{IFFT}\left(\text{Compose}(\mathbf{a}_{\text{norm}},  \mathbf{p}\right)).
\end{equation}

By combining the cross-attention mechanism and frequency domain analysis, our approach effectively mitigates the impact of false correlation on model performance while enhancing the model's ability to capture and comprehend visual and textual information directly relevant to the task.

\section{Experiment}

\subsection{Implementation Details}
\label{sec:implementation}

\noindent \textbf{Network Architecture.}\quad
Our network architecture is based on the FC-CLIP\cite{fc-clip} baseline, employing a ConvNeXt-Large CLIP backbone\cite{liu2022convnet} pretrained on the LAION-2B dataset\cite{schuhmann2022laion} using OpenCLIP. The pixel and mask decoders follow the default settings of Mask2Former\cite{ghiasi2022scaling} for generating category-independent masks. For in-vocabulary classification, we pool pixel features from the final decoder output based on mask predictions, as described in\cite{ghiasi2022scaling,fc-clip}, and compute the final classification logits via a matrix multiplication between the predicted class embeddings and the corresponding text embeddings of each class name.

\noindent \textbf{Training and Optimization.}\quad
For the training process, we follow the approach of \cite{cheng2022masked} and adopt the same training recipe and losses without any special design. The optimization is performed using the AdamW optimizer \cite{kingma2014adam} with a weight decay of 0.05. The input images are cropped to a size of $1024\times 1024$, and the initial learning rate is set to $1\times10^{-4}$ with a multi-step decay schedule employed to dynamically adjust the learning rate. The training batch size is 16, and the model is trained for 50 epochs on the COCO panoptic training set \cite{lin2014microsoft}.

\noindent \textbf{Inference Procedure.}\quad
At inference time, input images are resized such that the shorter side is $800$ pixels, while ensuring the longer side does not exceed $1333$ pixels. Mask predictions are merged using a mask-wise merging scheme\cite{cheng2022masked}. The out-vocabulary classifier operates on the frozen CLIP backbone features, and the final classification results are obtained through the geometric ensembling of in- and out-vocabulary classifiers\cite{ODISE,fc-clip}. We also incorporate prompt engineering methods from\cite{ODISE} and prompt templates from\cite{liang2023open} using their default settings.

\noindent \textbf{Evaluation Metrics and Datasets.}\quad
We primarily evaluate our model on the open-vocabulary semantic segmentation and open-vocabulary panoptic segmentation tasks. For open-vocabulary semantic segmentation, we perform zero-shot evaluation on the COCO\cite{lin2014microsoft}, ADE20K\cite{zhou2017scene}, and PASCAL\cite{everingham2010pascal} datasets. The open-vocabulary semantic segmentation results are evaluated using the mean Intersection-over-Union (mIoU) metric. For open-vocabulary panoptic segmentation, we evaluate the model on the COCO\cite{lin2014microsoft} and ADE20K\cite{zhou2017scene} datasets. We report the panoptic quality (PQ), semantic quality (SQ), and recognition quality (RQ) for open-vocabulary panoptic segmentation.
















\subsection{Main Results }
\label{sec:main_results}

\subsubsection{Open-vocabulary panoptic segmentation}

\begin{figure*}[!ht]
    \centering
    \includegraphics[width=15cm, height=7cm]{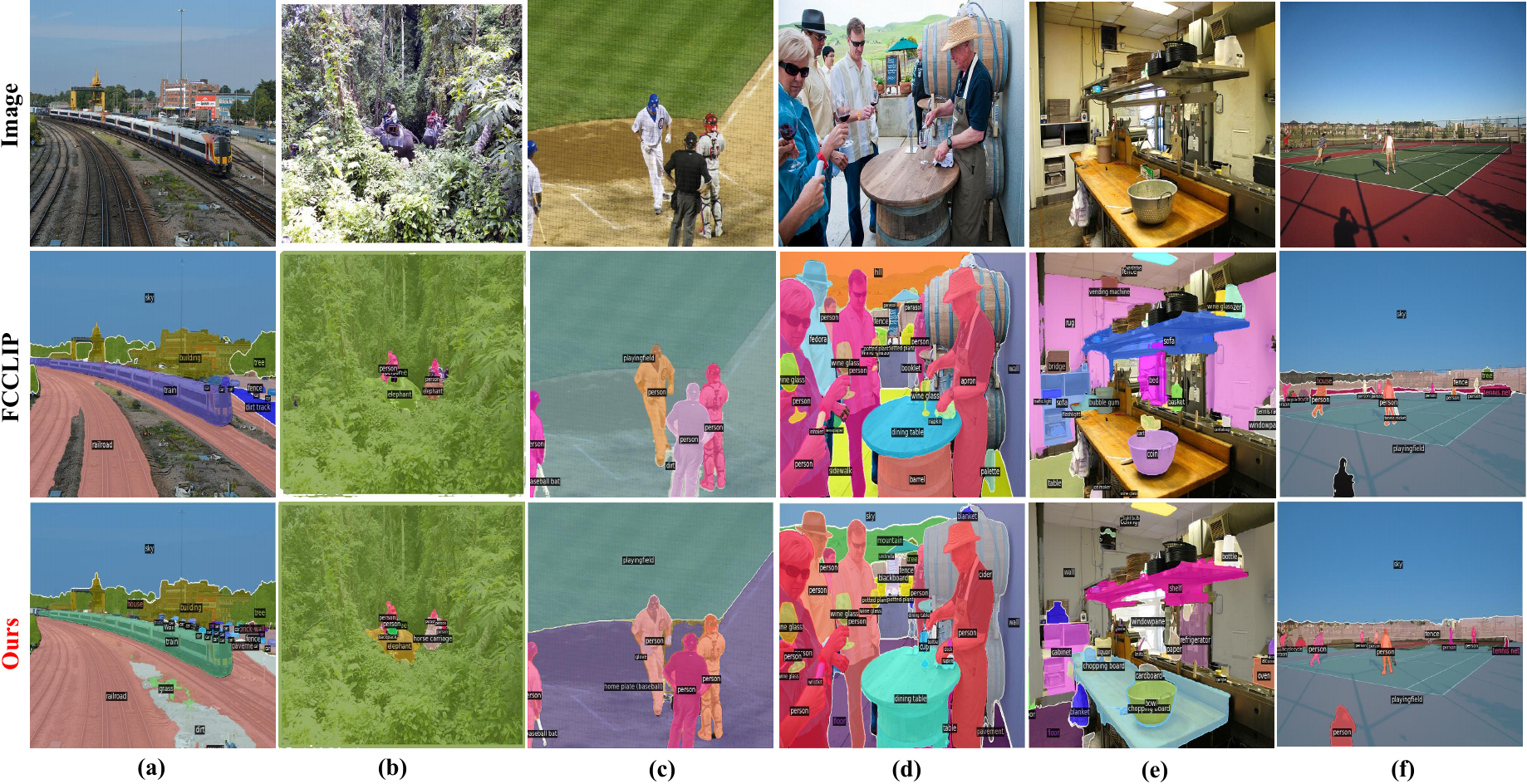}
    \caption{
\textbf{Qualitative Visualization of Open-Vocabulary Panoptic Segmentation.} 
To showcase the open-vocabulary recognition capability, we amalgamated class names from all datasets totaling approximately all classes, and conducted open-vocabulary inference directly. }
    \label{fig:pnopvis}
\end{figure*}

For open-vocabulary panoptic segmentation, we evaluate our method on the COCO\cite{lin2014microsoft} and ADE20K\cite{zhou2017scene} datasets, as shown in Table ~\ref{tab:ov_panoseg}. Compared to other methods, our approach demonstrates superior performance on both datasets. On the COCO dataset, we achieve a PQ of 57.5\%, surpassing the previous best method FC-CLIP\cite{fc-clip} by 3.1\%. Furthermore, our method yields an SQ of 83.7\% and an RQ of 67.9\%, outperforming FC-CLIP by 0.7\% and 3.1\%, respectively. These improvements indicate that our approach can effectively capture both the quality and completeness of panoptic segmentation masks.
On the more challenging ADE20K dataset, our method also sets a new state-of-the-art, achieving a PQ of 29.6\%, an SQ of 74.0\%, and an RQ of 35.8\%. Compared to the previous best method FC-CLIP, we obtain significant improvements of 2.8\%, 2.4\%, and 3.5\% in PQ, SQ, and RQ, respectively. These results demonstrate the robustness and generalization capability of our approach in handling diverse semantic categories and complex scenes.
In Figure \ref{fig:pnopvis}, we visualize the qualitative results of the unified open-vocabulary panoptic segmentation. It can be observed that our method not only recognizes more instances but also discovers more categories, such as "grass" and ‘dirty' in (a), and ‘glove' in (c). Furthermore, our method enhances the accuracy of object classification, as exemplified in (e).

\begin{table*}[t]
    \centering
    \caption{
        \textbf{Open-vocabulary panoptic segmentation performance.}
        ``COCO P.'' denotes the COCO panoptic datasets.
        ``COCO'' denotes the COCO image dataset. ``IN 1K'' denotes the ImageNet-1K image dataset.
        We report PQ, SQ and RQ for all datasets.
    }
    \label{tab:ov_panoseg}
    \renewcommand\arraystretch{0.9} {
    \setlength{\tabcolsep}{6mm}{
    \begin{tabular}{c|c|ccc|ccc}
    \toprule
    \hline
     &  & \multicolumn{3}{c|}{\textbf{COCO}} & \multicolumn{3}{c}{\textbf{ADE20K}} \\
     \cline{3-8}
     \multirow{-2}{*}{\textbf{Method}} & \multirow{-2}{*}{\textbf{Training Data}} & \textbf{PQ} & \textbf{SQ} & \textbf{RQ} & \textbf{PQ} & \textbf{SQ} & \textbf{RQ} \\
    \hline
    CutLER+STEGO\cite{wang2023cut} & IN 1K + COCO & 12.4 & 64.9 & 15.5 & - & - & - \\
    U2Seg\cite{niu2023unsupervised} & IN 1K + COCO & 16.1 & 71.1 & 19.9 & - & - & - \\
    MaskCLIP\cite{MaskCLIP} & COCO P. & {30.9} & {-} & {-} & 15.1 & 70.5 & 19.2 \\
    ODISE\cite{ODISE} & COCO P. & {55.4} & {-} & {-} & 22.6 & - & - \\
    FC-CLIP\cite{fc-clip} & COCO P. & {54.4} & {83.0} & {64.8} & 26.8 & 71.6 & 32.3 \\
    \rowcolor{gray}
    \textbf{Ours} &  COCO P. & \textbf{57.5} & \textbf{83.7} & \textbf{67.9} & \textbf{29.6} & \textbf{74.0} & \textbf{35.8} \\
    \hline
    \bottomrule
    \end{tabular}}}
\end{table*}

\subsubsection{Open-vocabulary semantic segmentation}

\begin{table*}[!ht]
    \centering
    \caption{
        \textbf{Open-vocabulary semantic segmentation performance.}
        We mainly compare with the fully-supervised and weakly-supervised methods.
        ``COCO S.'', ``COCO P.'' and ``COCO C.'' denote the COCO stuff, panoptic and caption datasets.
        ``O365'' denotes the Object 365 dataset. ``M. 41M'' denotes the merged 41M image dataset.
        We report mIoU for all datasets.
    }
    \label{tab:ov_semseg}
    \renewcommand\arraystretch{0.9} {
    \setlength{\tabcolsep}{4.5mm}{
    \begin{tabular}{l|c|cccccc}
    \toprule
    \hline
    \multirow{2}{*}{\textbf{Method}} & \multirow{2}{*}{\textbf{Training Data}} & \textbf{A-847} & \textbf{PC-459} & \textbf{A-150} & \textbf{PC-59} & \textbf{PAS-21} & \textbf{PAS-20} \\
    \cline{3-8}
     &  & \multicolumn{6}{c}{\textbf{mIoU (\%)}} \\
    \hline
    GroupViT\cite{xu2022groupvit}~$_{CVPR'22}$  & GCC + YFCC & 4.3 & 4.9 & 10.4 & 23.4 & 52.3 & 79.7 \\
    TCL\cite{cha2023learning}~$_{CVPR'23}$ & GCC & - & - & 14.9 & 30.3 & 51.2 & 77.5 \\
    OVSeg\cite{xu2023learning}~$_{CVPR'23}$ & CC4M & - & - & 5.6 & - & 53.8 & - \\
    SegCLIP\cite{luo2023segclip}~$_{ICML'23}$ & CC3M + COCO C. & - & - & 8.7 & - & 52.6 & - \\
    CLIPpy\cite{ranasinghe2023perceptual}~$_{ArXiv'23}$ & HQITP-134M & - & - & 13.5 & - & 52.2 & - \\
    MixReorg\cite{cai2023mixreorg}~$_{ICCV'23}$ & CC12M & - & - & 10.1 & 25.4 & 50.5 & - \\
    SAM-CLIP\cite{wang2023sam}~$_{ArXiv'23}$ & M. 41M & - & - & 17.1 & 29.2 & 60.6 & - \\
    SimBaseline\cite{SimSeg}~$_{ECCV'22}$  & COCO S. & - & - & 15.3 & - & 74.5 & - \\
    ZegFormer\cite{ding2022decoupling}~$_{CVPR'22}$ & COCO S. & - & - & 16.4 & - & 73.3 & - \\
    LSeg+\cite{li2022language}~$_{CVPR'23}$  & COCO S. & 3.8 & 7.8 & 18.0 & 46.5 & - & - \\
    X-Decoder\cite{zou2023generalized}~$_{CVPR'23}$ & COCO P. + C. & - & - & 25.0 & - & - & - \\
    OpenSEED\cite{zhang2023simple}~$_{ICCV'23}$ & COCO P. + O365 & - & - & 22.9 & - & - & - \\
    MaskCLIP\cite{MaskCLIP}~$_{ICML'23}$ & COCO P. & 8.2 & 10.0 & 23.7 & 45.9 & - & - \\
     OVSeg\cite{liang2023open}~$_{CVPR'23}$ & COCO S. & 9.0 & 12.4 & 29.6 & 55.7 & - & 94.5 \\
    SAN\cite{san}~$_{CVPR'23}$ & COCO S. & 13.7 & 17.1 & 33.3 & 60.2 & - & 95.5 \\
    OpenSeg\cite{ghiasi2022scaling}~$_{ECCV'22}$  & COCO P. + C. & 6.3 & 9.0 & 21.1 & 42.1 & - & - \\
    ODISE\cite{ODISE}~$_{CVPR'23}$   & COCO P. & 11.1 & 14.5 & 29.9 & 57.3 & 84.6 & - \\
    FC-CLIP\cite{fc-clip}~$_{NeurIPS'23}$ & COCO P. & 14.8 & 18.2 & 34.1 & 58.4 & 81.8 & 95.4 \\
    \hline
    \rowcolor{gray}
    \textbf{GBA (Ours)} &  COCO P.  &\textbf{15.1}            & \textbf{18.5}               & \textbf{35.9}                              & \textbf{59.6}      & \textbf{84.5}  & \textbf{95.8}     \\
    \hline
    \bottomrule
    \end{tabular}}}
\end{table*}

\begin{figure*}[!ht]
    \centering
    \includegraphics[width=15cm, height=7cm]{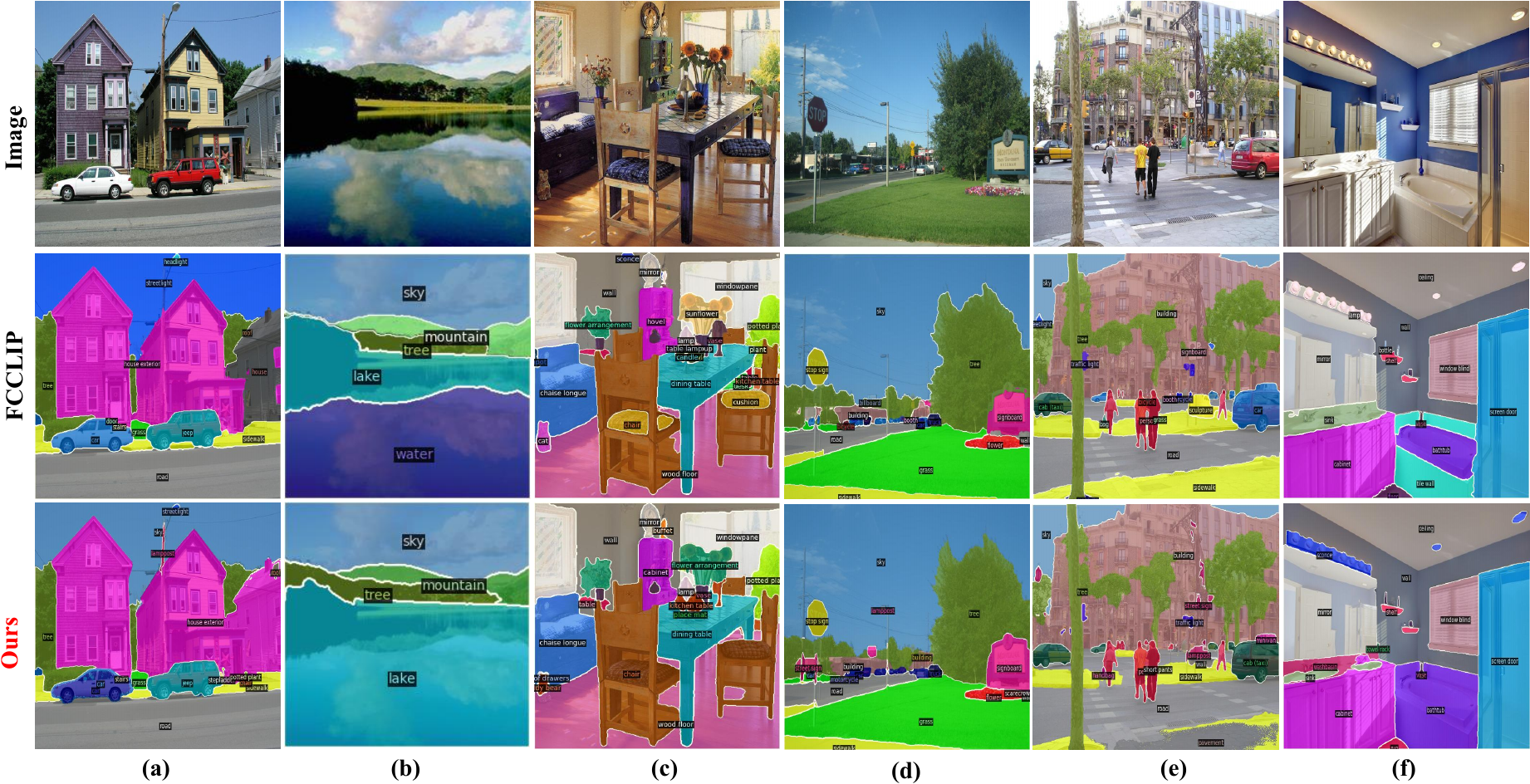}
    \caption{
 \textbf{Qualitative Visualization of Open-Vocabulary Semantic Segmentation}. To showcase the open-vocabulary recognition capability, we amalgamated class names from all datasets totaling approximately all classes, and conducted open-vocabulary inference directly. The second row represents the segmentation results of the baseline FC-CLIP, and the third row represents the segmentation results of our GBA.
    }
    \label{fig:semvis}
\end{figure*}

As shown in Table \ref{tab:ov_semseg}, we provide a comprehensive comparison of our GBA against prior works on a suite of benchmark datasets. These datasets include ADE20K\cite{zhou2017scene} (including 150 and 847 class variants), PASCAL Context\cite{mottaghi2014role} (459 and 59 class variants), and PASCAL VOC\cite{everingham2010pascal} (with 20 and 21 classes). GBA demonstrates consistent performance improvements across all evaluated datasets. Specifically, on the more challenging ADE20K-847 dataset, GBA achieves 15.1\% mIoU, outperforming the previous state-of-the-art method FC-CLIP\cite{fc-clip} by 0.3
\%. Similarly, on the PASCAL Context-459 dataset, GBA surpasses FC-CLIP by 0.3\%, reaching 18.5\% mIoU. These results highlight GBA's exceptional ability in classifying diverse semantic categories. Furthermore, on the PASCAL VOC\cite{everingham2010pascal} benchmark datasets, GBA exhibits substantial improvements, achieving 95.8\% and 84.5\% mIoU on the 20 and 21 class variants, respectively. This indicates that our GBA can accurately capture class distinctions and fine-grained spatial structures. 
In Figure \ref{fig:semvis}, we visualize the qualitative results of the unified open-vocabulary semantic segmentation. It can be observed from (b) that FC-CLIP\cite{fc-clip} predicts ‘lake' and ‘water' separately, while our method GBA accurately segments the entire ‘lake' region, demonstrating our method's superior ability to recognize and segment complete regions. As shown in (a), (c), and (d), our method also discovers more categories. Moreover, our method exhibits higher accuracy in object classification. For instance, in (c), we correctly identify the ‘bear' instead of misclassifying it as a ‘cat'. This can be attributed to our method's enhancement of feature diversity and suppression of spurious correlations.

\subsubsection{ADE20K Training and COCO Evaluation}
\begin{table}[!ht]
\centering
\setlength{\extrarowheight}{0pt}
\addtolength{\extrarowheight}{\aboverulesep}
\addtolength{\extrarowheight}{\belowrulesep}
\setlength{\aboverulesep}{0pt}
\setlength{\belowrulesep}{0pt}
\caption{\textbf{Results of training on ADE20K panoptic and evaluating on COCO panoptic val set.} The proposed GBA performs better than prior arts, even in the different setting (trained on ADE20K and zero-shot evaluated on COCO).}
\label{tab:train_with_ade}
\renewcommand\arraystretch{0.9} {
\begin{tabular}{l|ccc|ccc}
\toprule
& \multicolumn{3}{c|}{Zero-Shot Test Dataset}                & \multicolumn{3}{c}{Training Dataset}          \\
& \multicolumn{3}{c|}{COCO}                & \multicolumn{3}{c}{ADE20K}          \\
Method                    & PQ            & SQ            & RQ            & PQ            & SQ            & RQ            \\
\midrule \midrule
FreeSeg\cite{FreeSeg} & 16.5 & 72.0 & 21.6 & - & - & - \\
ODISE\cite{ODISE} & 25.0 & 79.4 & 30.4 & 31.4 & 77.9 & 36.9 \\
FC-CLIP\cite{fc-clip}  & 27.0 & 78.0 & 32.9 & 41.9 & 78.2 & 50.2 \\ 
\midrule
\textbf{GBA (Ours)}  & \textbf{28.2} & \textbf{78.8} & \textbf{34.1} & \textbf{42.8} & \textbf{80.5} & \textbf{51.3} \\ 
\bottomrule
\end{tabular}}

\end{table}

To further substantiate the efficacy of our proposed GBA approach, we conducted experiments utilizing a different training dataset. Specifically, following the methodology of \cite{ODISE}, we trained our model on the ADE20K dataset \cite{zhou2017scene} with panoptic annotations and evaluated its performance on the COCO panoptic dataset \cite{lin2014microsoft}. As illustrated in Table \ref{tab:train_with_ade}, even in this distinct setting (trained on ADE20K and zero-shot evaluated on COCO), our GBA method significantly outperformed the previous state-of-the-art methods, including FreeSeg \cite{FreeSeg}, ODISE \cite{ODISE}, and FC-CLIP \cite{fc-clip}, on the COCO dataset, achieving PQ improvements of 10.5, 2.0, and 2.0 percentage points, respectively.
Notably, although our model achieved a slightly lower semantic quality (SQ) score of 1.4 compared to ODISE \cite{ODISE}, which employs a significantly larger backbone network and thus a more robust mask generator, GBA still exhibited superior overall performance due to its simple yet effective design. This further corroborates the robustness and generalization capability of our approach, demonstrating its ability to maintain outstanding open-vocabulary segmentation performance even when applied to different datasets.

\begin{table}[!ht]
\centering
\setlength{\extrarowheight}{0pt}
\addtolength{\extrarowheight}{\aboverulesep}
\addtolength{\extrarowheight}{\belowrulesep}
\setlength{\aboverulesep}{0pt}
\setlength{\belowrulesep}{0pt}
\caption{\textbf{Results Of Variants.}
Experiments were conducted on PASCAL VOC 21, with mIoU as the evaluation metric. SDA+CCA refers to concatenating the two adapters with SDA first and CCA second, while CCA+SDA indicates the reverse order.}
\label{tab:var}
\renewcommand\arraystretch{0.9} {
\setlength{\tabcolsep}{4mm}{
\begin{tabular}{c|ccl} 
\toprule
Method    & Stage 1 & Stage 2 & Stage 3  \\ 
\hhline{====}
Plain  & 82.0   & 81.9    & 82.0    \\
SDA+CCA & 81.6    & 81.2    & 80.4     \\
CCA+SDA & 81.4    & 80.6    & 79.9     \\
\bottomrule
\end{tabular}}}
\end{table}

\subsubsection{Ablation Studies}

\begin{figure}[!ht]
    \centering
    \includegraphics[width=8.5cm, height=4.5cm]{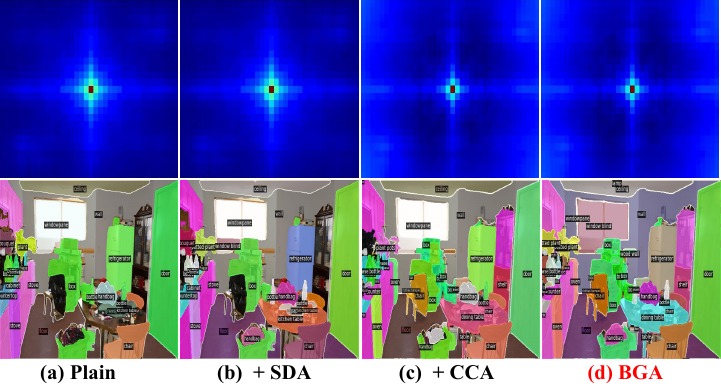}
    \caption{Visualization of the impact of Style Diversification Adapter (SDA) and Correlation Constraint Adapter (CCA) on frequency components. The changes in the heatmap of high and low-frequency components in the visual features output by CLIP (first row), and their influence on the final segmentation results (second row), after introducing the proposed SDA and CCA to the original CLIP model. }
  
    \label{fig:fft}
\end{figure}

We first investigate the impact of inserting vanilla adapters at different stages. As shown in Table \ref{tab:abl}, this approach slightly improves performance over the baseline FC-CLIP\cite{fc-clip} method at all stages, reaching a maximum mIoU of 82.0.
Incorporating SDA at the first stage significantly boosts performance, achieving an mIoU of 82.3. This finding suggests that SDA effectively enhances feature diversity in the early stages of the network, benefiting the subsequent learning process.
To independently assess the effectiveness of CCA, we conduct experiments without applying SDA. Inserting CCA at the third stage yields the best performance, reaching an mIoU of 82.5, indicating that CCA more effectively suppresses noisy information in the later stages of the network.
We also explore the joint impact of SDA and CCA. When SDA is applied at the first stage and CCA at the third stage, the most significant performance improvement is achieved, reaching an mIoU of 84.5. This result not only confirms the individual effectiveness of SDA and CCA but also highlights the advantages of their collaborative application at specific stages of the network.
Furthermore, we visualize the impact of SDA and CCA on the visual features output by CLIP. As shown in Figure\ref{fig:fft}, the frequency component heatmap of vanilla CLIP is primarily concentrated in the low-frequency regions. SDA enhances the style diversity of visual features through frequency-domain decomposition and amplitude modulation without affecting the spatial characteristics of high and low frequencies. After incorporating CCA, the features begin to focus significantly on high-frequency components. This indicates that CCA emphasizes high-frequency detail features relevant to textual semantics by leveraging cross-modal attention and frequency-domain normalization. These observations are consistent with our ablation study analysis presented in Table\ref{tab:abl}.

\subsection{Extending GBA to Transformer Architectures}

We apply two adapters from GBA (Style Diversification Adapter and Correlation Constraint Adapter) to the Transformer-based CLIP model, specifically the ViT-B/16 architecture used in ZegFormer\cite{ding2022decoupling} and ZegCLIP\cite{zegclip}.
We compare the enhanced versions with the original ZegFormer and ZegCLIP. Specifically, we strategically insert the two adapters into the 12 Transformer layers of CLIP: (1) SDA is inserted after the 4th layer to enhance style diversity in the early stages of feature extraction; (2) CCA is placed after the 8th layer to leverage high-level semantic features and suppress spurious correlations.
Experimental results (Table \ref{tab:transformer_results})demonstrate that even with this simple application, GBA achieves improvements on the Transformer architecture. On the PAS-21 dataset, the GBA-enhanced methods outperform the original ZegCLIP and ZegFormer in terms of mIoU for both seen and unseen categories. On the COCO-Stuff dataset, GBA-ZegFormer and GBA-ZegCLIP also exhibit moderate performance gains.
Considering that GBA was initially designed for CNN architectures, we believe that with comprehensive experimentation and optimization tailored to the Transformer architecture, GBA has the potential to achieve even greater performance improvements on Transformer-based models.

\begin{table}[ht]
\centering
\caption{\textbf{Comparison of Transformer-based methods on Pas-21 and COCO-Stuff 164K datasets.} Best results are in bold. These methods divide the classes in each dataset into seen (S) and unseen (U) categories, training only on seen classes and inferring on unseen classes.
}
\setlength{\tabcolsep}{0.8mm}{
\renewcommand\arraystretch{0.9} {
\begin{tabular}{l ccc ccc}
\toprule
\multirow{2}{*}{\textbf{Methods}} & \multicolumn{3}{c}{\textbf{PASCAL VOC 2012}} & \multicolumn{3}{c}{\textbf{COCO-Stuff 164K}} \\
\cmidrule(lr){2-4} \cmidrule(lr){5-7}
& \textbf{pAcc} & \textbf{mIoU(S)} & \textbf{mIoU(U)} & \textbf{pAcc} & \textbf{mIoU(S)} & \textbf{mIoU(U)} \\
\midrule
ZegFormer & -- & 86.4 & 63.6 & -- & 36.6 & 33.2 \\
\cellcolor[gray]{0.95}\textbf{+GBA} & \cellcolor[gray]{0.95}\textbf{--} & \cellcolor[gray]{0.95}\textbf{87.3} & \cellcolor[gray]{0.95}\textbf{64.2} & \cellcolor[gray]{0.95}\textbf{--} & \cellcolor[gray]{0.95}\textbf{37.3} & \cellcolor[gray]{0.95}\textbf{33.8} \\
ZegCLIP & 94.6 & 91.9 & 77.8 & 62.0 & 40.2 & 41.4 \\
\cellcolor[gray]{0.95}\textbf{+GBA} & \cellcolor[gray]{0.95}\textbf{96.7} & \cellcolor[gray]{0.95}\textbf{92.5} & \cellcolor[gray]{0.95}\textbf{78.8} & \cellcolor[gray]{0.95}\textbf{62.9} & \cellcolor[gray]{0.95}\textbf{40.6} & \cellcolor[gray]{0.95}\textbf{42.0} \\
\bottomrule
\end{tabular}}}

\label{tab:transformer_results}
\end{table}

\subsection{Exploration of variants of GBA}

Table~\ref{tab:var} shows the experimental results on variants of AIA. The results show that concatenating in this manner negatively impacts performance, as too many features are lost at the same stage, hindering the model's learning process. This effect is more pronounced in deeper stages.
Table~\ref{tab:var} presents the experimental results of different variants of the proposed method on the PASCAL VOC 21 dataset, using mean Intersection over Union (mIoU) as the evaluation metric. The plain model, without any adapters, serves as the baseline. 
Two variants, SDA+CCA and CCA+SDA, are investigated, where SDA and CCA adapters are concatenated in different orders.
The results demonstrate that concatenating adapters leads to a slight performance degradation compared to the plain model across all three stages. The SDA+CCA variant achieves mIoU scores of 81.6\%, 81.2\%, and 80.4\% in \textit{stages} 1, 2, and 3, respectively, while the CCA+SDA variant obtains 81.4\%, 80.6\%, and 79.9\% in the corresponding stages. This suggests that the concatenation of adapters may result in some information loss, which hinders the model's learning process. Notably, the performance drop becomes more pronounced in deeper stages, indicating that the impact of information loss accumulates as the network depth increases.

\begin{table}[!ht]
\centering
\caption{ \textbf{Ablation Studies.} \textit{Stage} 1–3 indicates the location where the adapter is inserted. The \CheckmarkBold symbol indicates the selected setting, while the \CheckmarkBold\kern-1.2ex\raisebox{1ex}{\rotatebox[origin=c]{125}{\textbf{--}}} symbol represents the opposite. 
\textit{w.} plain adapter denotes the replacement of SDA and CCA with plain adapters that does not employ the enhanced strategies.
}
\setlength{\tabcolsep}{1.1mm}{
\renewcommand\arraystretch{1} {
\begin{tabular}{l|ccc|c|c} 
\toprule[1px]
\textbf{Config.}                 & \textbf{Stage} 1 & \textbf{Stage} 2 & \textbf{Stage} 3 & {Method}                   & mIoU  \\ 
\hline\hline
Baseline                &         &         &         & FC-CLIP                  & 81.8  \\ 
\hline
\textit{w.} plain adapter        &   \CheckmarkBold       &         &         &                          & 82.0  \\
        &         &    \CheckmarkBold      &         &                          & 81.9  \\
        &         &         &      \CheckmarkBold    &                          & 82.0  \\
\cline{1-6}\cline{6-6}        
\multirow{3}{*}{\begin{tabular}[c]{@{}l@{}}Baseline\\+ SDA\end{tabular}} & \CheckmarkBold       &         &         & \multirow{17}{*}{\textbf{our GBA}} & \textbf{82.3}  \\
                        &         & \CheckmarkBold\kern-1.2ex\raisebox{1ex}{\rotatebox[origin=c]{125}{\textbf{--}}}        &         &                          & 82.1  \\
                        &         &         & \CheckmarkBold\kern-1.2ex\raisebox{1ex}{\rotatebox[origin=c]{125}{\textbf{--}}}        &                          & 81.9  \\ 
\cline{1-6}\cline{6-6}
\multirow{3}{*}{\begin{tabular}[c]{@{}l@{}}Baseline\\+ CCA\end{tabular}}  & \CheckmarkBold\kern-1.2ex\raisebox{1ex}{\rotatebox[origin=c]{125}{\textbf{--}}}        &         &         &                          & 82.5  \\
                        &         & \CheckmarkBold\kern-1.2ex\raisebox{1ex}{\rotatebox[origin=c]{125}{\textbf{--}}}       &         &                          & 82.9  \\
                        &         &         & \CheckmarkBold       &                          & \textbf{82.5}  \\ 
                        
\cline{1-6}\cline{6-6}

\multirow{3}{*}{\begin{tabular}[c]{@{}l@{}}Baseline\\+ SDA\\+ CCA\end{tabular}}  & \CheckmarkBold\kern-1.2ex\raisebox{1ex}{\rotatebox[origin=c]{125}{\textbf{--}}}        &         &         &                          & 83.3  \\
                        &         & \CheckmarkBold\kern-1.2ex\raisebox{1ex}{\rotatebox[origin=c]{125}{\textbf{--}}}       &         &                          & 83.8  \\
                        &         &         & \CheckmarkBold       &                          & \textbf{84.5}  \\ 
\cline{1-6}\cline{6-6}
\bottomrule[1px]
\end{tabular}
}}
\label{tab:abl}
\end{table}

\subsection{Efficiency Analysis}

We analyze the efficiency of our proposed method in terms of computational overhead, parameter count, and inference speed. As shown in Table~\ref{tab:model_complexity}, our method introduces a marginal increase of 3.23\% in total parameters and 1.42\% in GFLOPs compared to the baseline, indicating minimal computational overhead. Although the number of trainable parameters increases by 58.09\%, it primarily originates from the lightweight adapter module and constitutes a small fraction of the overall model size, resulting in a negligible increase in actual inference time.
Table~\ref{tab:fps_comp} presents a frames per second (FPS) comparison, where our method maintains an inference speed comparable to FC-CLIP and significantly outperforms ODISE\cite{ODISE}. On the COCO\cite{lin2014microsoft} and Pascal VOC 21\cite{everingham2010pascal} datasets, our approach achieves an FPS of 2.96 and 6.33, respectively.
Furthermore, we compare the parameter efficiency and computational cost of our method with other approaches in Table~\ref{tab:adapter_comparison}. Despite using a higher input resolution and introducing additional parameters through the adapter module, our approach achieves competitive performance while maintaining a reasonable computational cost.

\begin{table}[ht]
\centering
\caption{\textbf{Comparison of model complexity.}}
\setlength{\tabcolsep}{0.9mm}{
\renewcommand\arraystretch{0.9} {
\begin{tabular}{lrrc}
\hline
\textbf{Metric} & \textbf{Baseline} & \textbf{Our Method} & \textbf{Increase (\%)} \\
\hline
Total Params & 372,494,210 & 384,530,690 & 3.23\% \\
Trainable Params & 20,721,601 & 32,758,081 & 58.09\% \\

\hline
\end{tabular}}}

\label{tab:model_complexity}
\end{table}

\begin{table}[ht]
\centering

\caption{\textbf{FPS comparison.} FPS comparison obtained using one A6000. }
\label{tab:2}
\setlength{\tabcolsep}{0.8mm}{
\renewcommand\arraystretch{0.9} {
\begin{tabular}{c|cc} 
\toprule
Method                                  & COCO & Pascal VOC 21 \\ 
\hline\hline
ODISE & 0.52 & 0.41          \\
FC-CLIP                                 & 3.15 & 6.45          \\
\hline
\textbf{GBA (Ours)}                                & \textbf{2.96} & \textbf{6.33}          \\
\bottomrule
\end{tabular}}}
\label{tab:fps_comp}
\end{table}

\begin{table}[htbp]
\centering
\caption{\textbf{Comparison of parameter efficiency and computational cost across different methods. 
'Adapter P.' shows the number of parameters introduced by the adapter module (if applicable). 'Trainable P.' represents the total number of trainable parameters in millions. 
Our method uses ConvNeXt as the backbone with an input resolution of 1024$\times$1024, while other methods use ViT-B/16 with 640$\times $640 input resolution. OVSeg}~\cite{liang2023open}has a similar structure to SimSeg~\cite{SimSeg} but finetunes the entire CLIP model, resulting in significantly more trainable parameters. '--' denotes unavailable data.}
\label{tab:adapter_comparison}
\setlength{\tabcolsep}{0.9mm}{
\renewcommand\arraystretch{0.9} {
\begin{tabular}{lcccc}
\hline
\textbf{Method} & \textbf{Backbone} & \textbf{Adapter P.} & \textbf{Trainable P.} & \textbf{GFLOPs} \\
\hline
SAN\cite{san} & ViT-B/16 & 8.4 M & 8.4 M & 64.3 \\
MaskCLIP\cite{ding2022open} & ViT-B/16 & -- & 63.1 M & 307.8 \\
SimSeg~\cite{SimSeg} & ViT-B/16 & -- & 61.1 M & 1916.7 \\
OvSeg\cite{liang2023open} & ViT-B/16 & -- & 147.2 M & 1916.7 \\
\hline
FC-CLIP\cite{fc-clip} & ConvNeXt-L & -- & 19.8 M & 986.1 \\
\cellcolor[gray]{0.95}\textbf{BGA (Ours)} &\cellcolor[gray]{0.95} ConvNeXt-L & \cellcolor[gray]{0.95}11.5 M & \cellcolor[gray]{0.95}31.3 M & \cellcolor[gray]{0.95}1000.2 \\
\hline
\end{tabular}}}
\end{table}

\subsection{Visualization Assessment on More Challenging Datasets}

\begin{figure*}[!ht]
    \centering
    \includegraphics[width=16cm, height=7cm]{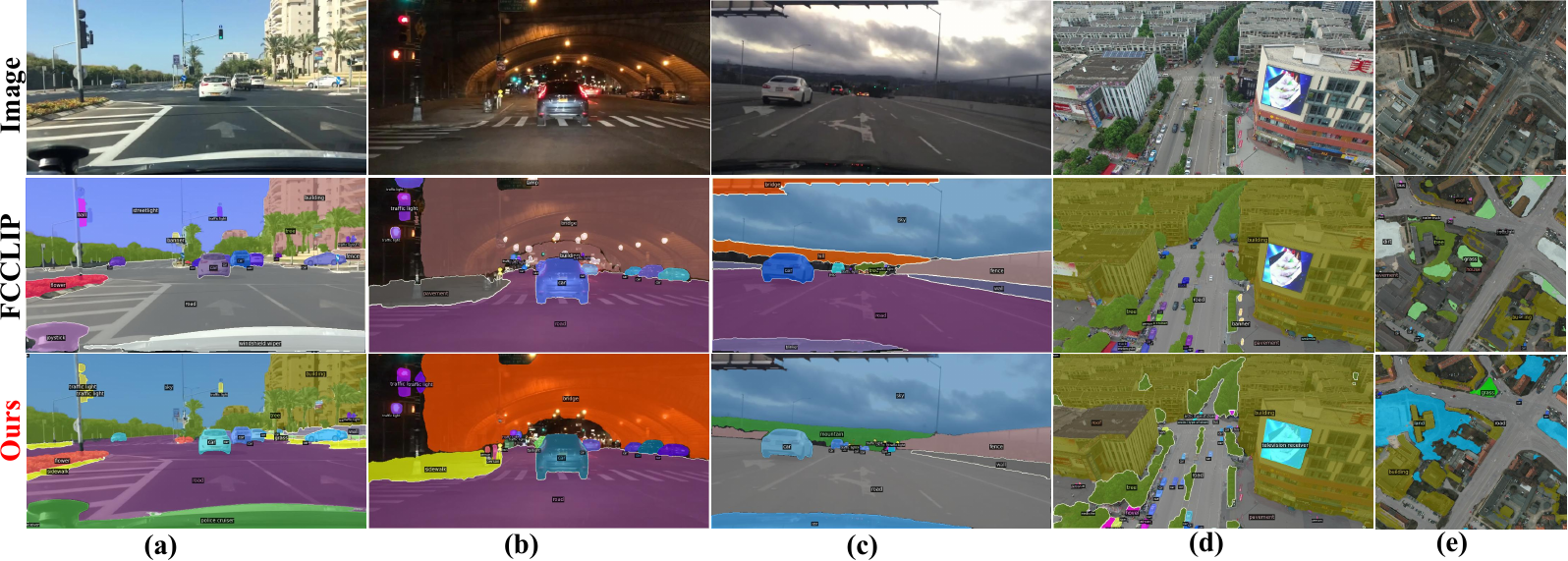}
    \caption{Open-vocabulary panoptic segmentation performance on the BDD-100K ,UAVid and ISPRS Potsdam datasets. (a), (b) and (c) represent sunny, nighttime, and cloudy scenes, respectively, which represent different image styles and showcase the domain generalization ability of our BGA. }
    \label{fig:res_vis}
\end{figure*}

To further assess the performance of our method on more challenging images, we have selected three representative datasets from the MESS benchmark\cite{mess}: the driving imagery dataset BDD-100K\cite{bdd100k}, the aerial imagery dataset UAvid, and the remote sensing imagery dataset ISPRS Potsdam\cite{potsdam}.
Experimental results on BDD-100K (Figure \ref{fig:res_vis} (a),(b) and (c)) demonstrate that GBA exhibits stronger generalization ability on images with different styles compared to the baseline methods. Benefiting from the robustness of the SDA module to image style variations and the enhancement of semantic associations by the CCA module, GBA can more accurately segment target regions under various lighting conditions and effectively suppress background interference.
However, the performance on the UAvid\cite{uavid} and ISPRS Potsdam\cite{potsdam} datasets reveals the challenges faced by cross-domain generalization (Figure\ref{fig:res_vis} (d) and (e)). Aerial and remote sensing images differ significantly from the COCO dataset used for fine-tuning in terms of shooting angle, target scale, and texture features, leading to a larger domain gap between the source dataset and the target domain, which may limit the performance improvement of our method in these specific domains.
Despite the challenges in cross-domain generalization, our GBA method demonstrates strong generalization ability on datasets closer to natural images (such as BDD-100K\cite{uavid}), proving the effectiveness of SDA and CCA in handling the diversity and complexity of natural scenes.

\section{Limitations}

We propose a novel single-stage open-vocabulary segmentation framework that achieves state-of-the-art performance through simple yet effective adapter structures. Despite the encouraging results, we recognize that there are still some limitations and challenges that require further investigation.
The main limitations include the need to improve segmentation accuracy for occluded instances and the need to enhance the framework's ability to identify camouflaged instances. These challenges highlight the necessity for further research to improve the robustness and generalization capability of the framework.
Furthermore, we identify two interesting future research directions. First, exploring methods to decouple spurious correlations from text features is crucial to prevent the model from being misled by irrelevant information. Second, developing effective techniques to handle conflicting or overlapping vocabulary items, such as distinguishing semantically related but hierarchically different entities (e.g., "dog" and "dog tail"), is an important direction for future research.

\section{Conclusion}

The proposed Generalization Boosted Adapter (GBA) strategy effectively enhances the generalization capability and robustness of open-vocabulary segmentation tasks based on the CLIP model. GBA consists of two key components: Style Diversification Adapter (SDA) and Correlation Constraint Adapter (CCA). The SDA, acting on the amplitude component, enriches the feature space representation while preserving content information, thereby mitigating the overfitting problem caused by limited fine-tuning data. The CCA, introduced into the deep layers of the visual encoder, suppresses low-frequency "noise" and improves the accuracy of category matching by avoiding erroneous associations between the correct category and irrelevant features. The synergistic effect between the shallow SDA and deep CCA enables GBA to effectively alleviate overfitting and enhance the semantic relevance of feature representations. Extensive experiments validate the effectiveness of GBA, demonstrating state-of-the-art performance on multiple benchmarks. The results highlight the potential of GBA as a general solution for improving the generalization capability and robustness of CLIP-based models, paving the way for the design of novel adapter modules in various computer vision tasks.

\bibliographystyle{IEEEtran}
\bibliography{ref}

\end{document}